\documentclass[runningheads]{llncs}

\usepackage{eccv}

\usepackage{eccvabbrv}

\usepackage{graphicx}
\usepackage{booktabs}
\usepackage{amsmath,amssymb}
\usepackage{algorithm}
\usepackage{algpseudocode}
\usepackage{multirow}
\usepackage{epigraph}
\usepackage[table,dvipsnames]{xcolor}
\usepackage{tcolorbox}
\usepackage{adjustbox}
\usepackage{enumitem}

\usepackage[accsupp]{axessibility}  

\usepackage[pagebackref,breaklinks,colorlinks,citecolor=eccvblue]{hyperref}

\usepackage{orcidlink}

\begin{document}


\newcommand{\red}[1]{{\color{red}#1}}
\newcommand{\todo}[1]{{\color{red}#1}}
\newcommand{\TODO}[1]{\textbf{\color{red}[TODO: #1]}}
\newcommand{\tocite}{\textcolor{red}{[TOCITE]~}}

\newcommand{\zhy}[1]{\textcolor{orange}{ZP: #1}}
\newcommand{\parahead}[1]{\noindent\textbf{#1}.\ }
\newcommand{\brian}[1]{{\color{magenta} brian: #1}}
\newcommand{\ew}[1]{{\color{teal}[EW: #1]}}

\newcommand{\datasetname}{AiM3D\xspace}
\newcommand{\methodname}{Kirin\xspace}

\definecolor{tabfirst}{rgb}{1, 0.2, 0.2} 
\definecolor{tabsecond}{rgb}{0.2, 0.5, 1} 
\definecolor{tabthird}{rgb}{1, 1, 0.7} 

\newcommand{\et}[2]{${#1}_{\pm{#2}}$}
\newcommand{\etb}[2]{$\mathbf{{#1}}_{\pm{#2}}$}
\newcommand{\etfirst}[2]{$\textcolor{tabfirst}{{\textbf{#1}}}_{\pm{#2}}$}
\newcommand{\etsecond}[2]{$\textcolor{tabsecond}{\textbf{{#1}}}_{\pm{#2}}$}
\newcommand{\ets}[2]{$\underline{{#1}}_{\pm{#2}}$}

\setlist[itemize]{noitemsep, topsep=0pt}

\makeatletter
\newcommand{\printfnsymbol}[1]{%
  \textsuperscript{\@fnsymbol{#1}}%
}
\makeatother








\title{
    \methodname: Animal Motion Generation from In-the-Wild Video
} 


\author{
Brian Nlong Zhao \inst{1}\thanks{Equal contribution.} \and
Zhuoyang Pan \inst{2}\printfnsymbol{1} \and \\
James M. Rehg \inst{1} \and
Jiajun Wu \inst{3} \and
Shangzhe Wu \inst{4}
}

\authorrunning{BN.~Zhao et al.}

\institute{
University of Illinois Urbana-Champaign \and
University of Pennsylvania \and
Stanford University \and
University of Cambridge
}

\maketitle

\begin{center}
    \vspace{-0.5em}
    \captionsetup{type=figure}
    \includegraphics[width=0.85\linewidth]{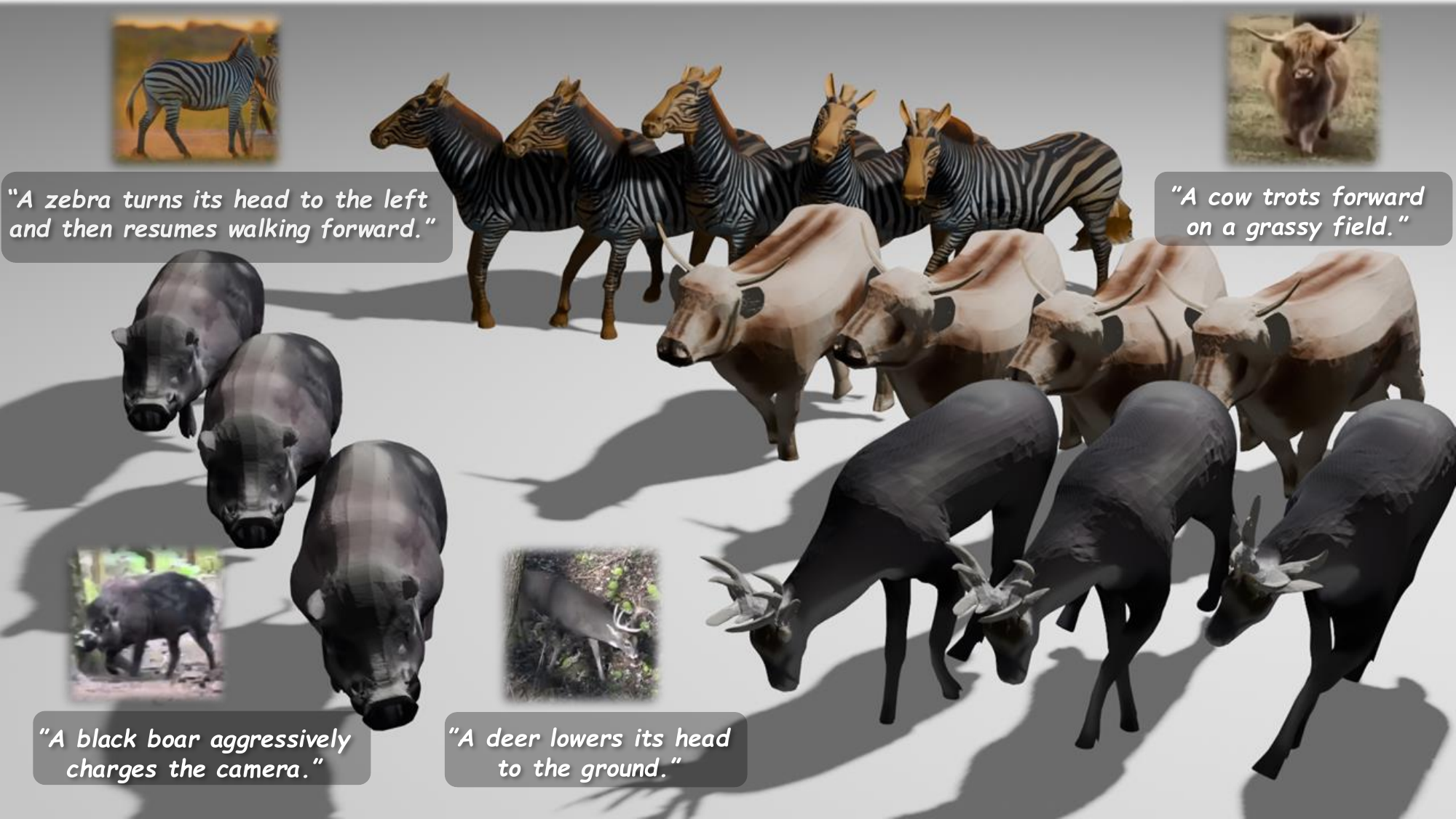}
    \vspace{-0.5em}
    \captionof{figure}{
    Our framework takes an animal image and a text description of the desired motion as input, and generates a 3D animated mesh sequence of the animal.
    This is achieved by a animal motion generation model trained using 3D motion sequences extracted from large-scale in-the-wild video, in conjunction with an automatic rigging process to produce a fully animated mesh.
    }
    \label{fig:teaser}
\end{center}

\begin{abstract}
Understanding animal motion is fundamental to modeling animal behavior and biomechanics, yet progress in this area lags far behind human motion research due to the scarcity of high-quality motion data. While human motion can be captured in controlled environments, it is impractical for most animal species, resulting in small, domain-limited datasets that restrict downstream applications such as animation. To address this challenge, we introduce \methodname, a framework that reconstructs motion from video, learns motion priors at scale, and generates realistic motion that can be directly applied to animated assets. Using large collections of in-the-wild animal videos, we reconstruct 3D motion sequences and pair them with captions to create \datasetname, the first large-scale dataset offering aligned video-text-motion tuples for quadruped animals. Building on this dataset, we develop a visual-guided motion generation model that conditions on both text and image to guide the generation of realistic motion across diverse animal species. Finally, by leveraging an off-the-shelf image-to-3D model, we automatically rig and animate 3D meshes using generated motion, producing ready-to-render animated animals. Together, our dataset and framework establish a new foundation for large-scale, text and image conditioned animal motion generation and animation. Project page: \url{https://kirin-ani.github.io/}.
\end{abstract}

\section{Introduction}
\label{sec:introduction}

\setlength{\epigraphwidth}{.85\textwidth}
\epigraph{\emph{“Elsewhere we have investigated in detail the movement of animals… there remains an investigation of the common ground of any sort of animal movement whatsoever.”}
}{\textsc{Aristotle, On the Motion of Animals}}

Understanding how animals move in their natural habitats is a fundamental scientific challenge, with implications for studying behavior in biology and ecology, and for building more generalizable motion models in computer vision.
Yet, compared to human motion, animal motion modeling remains vastly underexplored, primarily due to the lack of large-scale, high-quality motion data.
For humans, motion capture technologies have led to large-scale precise 3D motion datasets, enabling rapid progress in 3D human modeling, animation, and generation. However, it is challenging to capture animal motion in controlled laboratory settings at scale, and impractical for wild or endangered species. This data scarcity has become the main bottleneck preventing progress in animal motion research.

One common solution is to rely on human crafted motion data. Previous efforts such as DeformingThings4D \cite{li20214dcomplete} and Truebones Zoo \cite{truebones} include animal motions manually created by computer graphics artists, while AniMo \cite{wang2025animo} extracts motion sequences from video games that are similarly authored by human designers and animators. Although these datasets provide high quality and physically consistent motion, they are inherently constrained to a limited set of predefined actions such as walking, eating, or sleeping, and therefore fail to capture the full diversity and natural variability of animal behaviors observed in the wild.

On the other hand, the Internet offers an abundance of in-the-wild animal footage capturing diverse species, behaviors, and environments, far beyond what controlled datasets can provide. Meanwhile, recent advances in 3D reconstruction and shape estimation from monocular images and videos have made it possible to recover 3D structures from unstructured video data. Together, these developments open a new opportunity: \emph{Can we learn realistic, generalizable models of animal motion directly from in-the-wild videos?}


Several prior works have explored this direction and made notable progress. Ponymation~\cite{sun2024ponymation} extends MagicPony \cite{wu2023magicpony}~to learn an \emph{unconditional} generative model of 3D motion from videos of a single horse category. AiM \cite{animal-in-motion} presents a large scale dataset of animal videos from the Internet spanning 23 quadruped categories,
but does not attempt to learn a generative model of their underlying motions.

In this paper, we present \methodname, a comprehensive framework for learning and generating 3D quadruped motion directly from large-scale video data.
Our pipeline begins by enhancing existing 3D reconstruction methods to recover accurate and temporally smooth motion sequences from videos. Leveraging the AiM dataset, 
%
%
we develop a SMAL-based \cite{zuffi20173dsmal} 3D motion reconstruction system that produces consistent and realistic motion trajectories. Each reconstructed sequence is further paired with descriptive textual annotations generated by VLMs, resulting in the first large-scale dataset containing aligned text–video–motion tuples for quadruped animals.

To fully exploit the visual and semantic cues in this dataset, we introduce a novel adaptation of MDM \cite{tevet2023humanmdm}, which yields an animal motion generation model capable of conditioning on both text and visual input. Unlike prior approaches that rely on manually designed or synthetic motion data, our method learns directly from in-the-wild videos, capturing a broader and more diverse spectrum of natural animal motion. Experimental results show that our dataset and generation model achieve state-of-the-art performance on both our test set
and external out-of-distribution test sets, establishing a scalable foundation for data-driven animal motion modeling.

Furthermore, by leveraging off-the-shelf image-to-3D generation tools \cite{zhang2024clay}, \methodname can automatically rig the generated 3D mesh and apply the generated motion to produce realistic animated 3D mesh sequences. Comparisons show that our animation method produces more plausible 3D motion sequences compared to baseline approaches, while being more efficient. In summary, our work's contributions are as follows:
\begin{itemize}
    \item We enhance state-of-the-art 3D animal pose reconstruction methods for video-based reconstruction and create \datasetname dataset, a large-scale animal motion dataset with aligned text, video, and motion data.
    \item We propose the first animal motion generation model, conditioning on both text and image input.
    \item Experiments demonstrate that training on our dataset with visual conditioning achieves state-of-the-art results on both in-distribution and external out-of-distribution test sets, highlighting the effectiveness of our dataset and model.
    \item In conjunction with a text-to-3D model, we present a fully automatic system, \methodname, that turns a 2D image into a realistic 3D animated mesh sequence, outperforming existing 4D animation methods.
\end{itemize}

\begin{figure*}[!htpb]
    \centering
    \includegraphics[width=\linewidth]{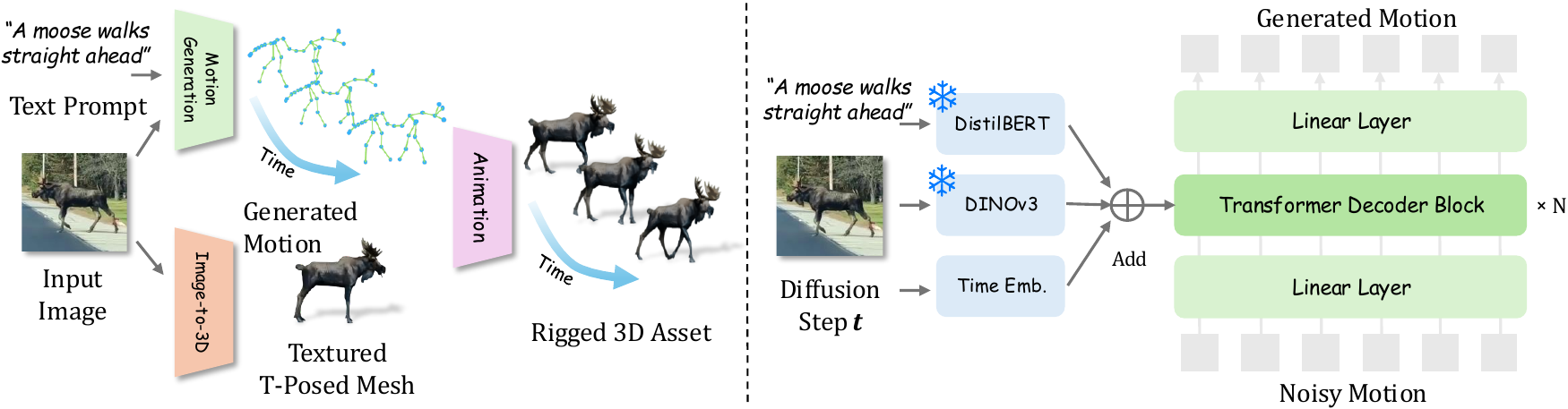}
    \vspace{-2em}
    \caption{
    \textbf{Left: Overview of \methodname generation pipeline.} A text description and an image are provided as inputs. The text and image are used for motion generation, while the image is also used to generate a T-posed mesh. The animation module then rigs the generated motion onto the mesh to produce the final animated 3D model. \textbf{Right: Overview of the motion generation architecture.} Text features are extracted using a frozen DistilBERT encoder, and image features are extracted using a frozen DINOv3 encoder. The text, image, and denoising step embeddings are combined and fed into a transformer decoder with cross-attention to generate motion sequences.
    }
    \vspace{-3mm}
    \label{fig:method}
\end{figure*}

\section{Related Works}
\label{sec:related_works}

\subsection{Animal Motion Datasets}
\label{sec:animal-motion-datasets}

A major obstacle in animal motion generation is the lack of high quality motion data. Unlike humans, whose movements can be captured in controlled laboratory environments, most animal species cannot be brought into motion capture facilities and rarely behave naturally under such conditions. Their wide variety of anatomies and behaviors also makes standardized capture difficult.

Existing motion capture datasets collected in controlled settings \cite{kearney2020rgbd, shooter2024benchmarking, varen} cover only a few species and offer limited ecological validity. Other efforts rely on artist created or human adapted motion data \cite{truebones, li20214dcomplete, wang2025animo, yang2023omnimotiongpt}, which are expensive to produce and introduce a significant domain gap.

Learning animal motion directly from videos is promising, but current data remains insufficient. BADJA \cite{biggs2018creaturesbadja} has only 11 sequences with 3D pose. AnimalKingdom \cite{animalkingdom} provides single image pose without motion. COP3D \cite{sinha2022commoncop3d} contains mostly orbit shots of stationary pets and is limited to common categories such as cats and dogs. APT-36K \cite{yang2022apt} offers only fifteen frame clips, providing limited motion. Even the largest dataset, AiM \cite{animal-in-motion}, contains nearly 30k videos but still lacks 3D motion.

To address these limitations, we build upon AiM by reconstructing SMAL based 3D pose from video to obtain high quality motion sequences. We also use a vision language model to generate multiple behavior aware text descriptions for each clip, resulting in about 30k motion sequences paired with about 180k captions. To the best of our knowledge, this is the first large scale animal motion dataset with aligned video, motion, and descriptions for diverse animal species.

\subsection{3D Animal Reconstruction}
Many works estimate 3D or 4D animal shape and pose from single images or videos, using either model-based or model-free approaches, and either feed-forward or optimization-based pipelines.

Model-based methods such as ABM \cite{badger20203dabm}, SMAL \cite{zuffi20173dsmal}, and their extensions \cite{rueegg2022barc,bite2023rueegg,biggs2018creaturesbadja,biggs2020leftsmalify,liontigerbear,zuffi2019threedsafari,li2021coarse,wang2021birds,baieri2025modeldolphin,zhong20254danimal} optimize a parameterized mesh to fit 2D labels such as masks or keypoints. These optimization-based methods often overfit to 2D projections and may yield unnatural 3D geometry. More recent systems \cite{lyu2025animer,sabathier2024animalavatar} leverage SMAL to create image–3D paired data and enable feed-forward inference, reducing 3D inconsistencies with only minor loss in per-frame projection accuracy.

Model-free methods \cite{wu2023magicpony,safari_from_visual_signals,de20244dpv,kanazawa2018learningcmr,kulkarni2019canonicalcsm,kulkarni2020articulationacsm,li2020selfunsupmesh,wu2023dove,yao2022lassie} learn shape directly from large datasets and provide flexible feed-forward predictions, but typically lack explicit skeletal structures, which limits their suitability for reconstructing articulated motion. Similarly, generic 3D and 4D reconstruction frameworks \cite{ren-redo2021,ren2024l4gm,yang2022banmo,yang2021viser} can recover animal shape but do not supply consistent skeleton definitions.

Given these limitations, and since our goal is accurate skeletal motion rather than perfect shape, we adopt AniMer \cite{lyu2025animer} as initialization, which uses SMAL model trained on 3D dataset, providing universal skeleton and at the same time avoiding unnatural 3D poses.

\subsection{Animal Motion Generation}

Although large scale data for animal motion generation remains limited, several recent works have begun exploring this direction. OmniMotionGPT \cite{yang2023omnimotiongpt} compensates for data scarcity by combining human motion prior with small human crafted animal motion datasets to transfer motion knowledge. AniMo \cite{wang2025animo} trains a two stage RVQ \cite{lee2022autoregressivervq} model on artist made motion sequences extracted from video games to produce plausible synthetic motion.

Other efforts target different settings. SinMDM \cite{raab2024singlesinmdm} learns motion motifs from a single motion example using a diffusion model with a restricted receptive field, but it does not generalize beyond the given exemplar. Puppeteer \cite{song2025puppeteer} and MotionAvatar \cite{zhang2024motionavatar} animate auto rigged meshes by matching or applying generated motion, yet both rely on synthetic motion sources rather than real world video.

The closest work to ours is Ponymation \cite{sun2024ponymation}, which collects online horse videos and reconstructs 3D motion using the MagicPony \cite{wu2023magicpony} pipeline. However, their VAE based \cite{kingma2013autovae} model is unconditional and limited to a single species. In contrast, our work builds on Animal in Motion \cite{animal-in-motion} to construct a large scale dataset spanning twenty three quadruped categories with paired videos and descriptive text. While Ponymation uses images only for textured mesh inference, our model conditions motion generation directly on both text and images. We adopt MDM \cite{tevet2023humanmdm} as our backbone and extend it with an image conditioning branch to learn motion synthesis grounded in visual context.

\section{Method}
\label{sec:method}
Our method, \methodname, consists of three components: (1) given an animal video dataset, we first reconstruct motion from each clip to build a large-scale animal motion dataset; (2) using this dataset, we train a motion generation model; and (3) leveraging an image-to-3D module and an auto-rigging pipeline, we generate an animated 3D mesh based on user-provided text and image inputs.
See \cref{fig:method} for our generation framework overview.

\subsection{Extracting Animal Motions from Web Videos}
Our goal is to recover accurate, temporally coherent 4D animal motions from \emph{in-the-wild} web videos. To improve optimization stability and reconstruction quality, we decouple global translation estimation from local pose reconstruction, which allows the model to focus on optimizing fine-grained articulated motion without being affected by global displacement or noisy motion cues present in uncontrolled video data. Starting from a large collection of animal clips in the AiM dataset \cite{animal-in-motion}, we re-estimate per-video 3D body pose using a new SMAL-templated pipeline designed for robust projection alignment and efficient refinement.
SMAL is a general parametric template for quadrupeds with \emph{shape} coefficients $\boldsymbol\beta\in\mathbb{R}^{41}$, \emph{pose} parameters $\boldsymbol\theta\in\mathbb{R}^{35\times 3}$, and a skinned mesh articulated by $J{=}35$ joints.
We leverage AniMer~\cite{lyu2025animer} for per-frame initialization and perform sequence-level refinement to enforce tighter keypoint alignment and temporal smoothness.
Global translation is estimated separately using SpatialTrackerV2 \cite{xiao2025spatialtracker}, an off-the-shelf 3D point tracking model, and is later combined with the optimized pose sequence to recover the complete global 4D motion.

\parahead{Initialization}
Each video in the dataset contains a single animal per frame by construction. For each frame $\mathbf{I}_t$, we use precomputed instance masks $\mathbf{M}_t$ from Grounded-SAM2 \cite{ravi2024sam2segmentimages,ren2024grounded} and 2D keypoints $\mathbf{P}_t$ from ViTPose++ \cite{xu2023vitpose++}. We obtain per-frame initial estimates using Animer \cite{lyu2025animer}:
\(\{(\boldsymbol\beta_t^{(0)},\,\boldsymbol\theta_t^{(0)},\,\boldsymbol\pi_t^{(0)})\}_{t=1}^T\),
with \(\boldsymbol\pi_t^{(0)}=(\mathbf{K},\mathbf{E}_t)\), where:
(i) $\boldsymbol\beta_t^{(0)}\!\in\!\mathbb{R}^{41}$ are shape coefficients of SMAL,
(ii) $\boldsymbol\theta_t^{(0)}\!\in\!\mathbb{R}^{35\times3}$ are joint poses in axis–angle (35 joints),
and (iii) $\boldsymbol\pi_t^{(0)}=(\mathbf{K},\mathbf{E}_t)$ is a weak-perspective camera with fixed intrinsics
$\mathbf{K}=\mathrm{diag}(f,f,1)$ ($f{=}1000$) and translation $\mathbf{E}_t\!\in\!\mathbb{R}^3$.
We initialize a sequence-shared shape $\boldsymbol\beta^{(0)}=\frac{1}{T}\sum_t \boldsymbol\beta_t^{(0)}$,
convert poses to 6D $\mathbf{r}_t^{(0)}\!\in\!\mathbb{R}^{35\times6}$, and keep cameras fixed. 

\parahead{Objective}
We optimize the sequence-shared shape $\boldsymbol{\beta}$ and the per-frame
6D joint rotations $\{\mathbf r_t\}_{t=1}^T$, while keeping the cameras fixed.
We convert the 6D rotations to rotation matrices
$\mathbf R_t \in \mathrm{SO}(3)^{35}$ using the standard 6D-to-SO(3) mapping
$\rho(\cdot)$, where
$
    \mathbf R_t = \rho(\mathbf r_t).
$
The optimization objective is:
\begin{equation}
\begin{aligned}
\min_{\boldsymbol\beta,\{\mathbf r_t\}} \ \mathcal{L}
&= \lambda_{\text{proj}} \mathcal{L}_{\text{proj}}
 + \lambda_{\text{smooth}} \mathcal{L}_{\text{smooth}} \\
\mathcal{L}_{\text{proj}}
&= \sum_{t=1}^{T}\frac{\big\|(\hat{\mathbf{k}}_t-\mathbf{k}_t)\odot \mathbf{w}_t\big\|_1}{\sum \mathbf{w}_t + \varepsilon}, \\
\mathcal{L}_{\text{smooth}}
&= \sum_{t=2}^{T} d_{\mathrm{SO(3)}}^2(\mathbf{R}_{t-1},\mathbf{R}_t) \\
& + \alpha \sum_{t=3}^{T} d_{\mathrm{SO(3)}}^2(\mathbf{R}_{t-1}^\top\mathbf{R}_t,\mathbf{R}_{t-2}^\top\mathbf{R}_{t-1}). \\
\end{aligned}
\end{equation}
Here $\hat{\mathbf{k}}_t$ are SMAL keypoint projections aligned to 2D keypoint indices, $\mathbf{w}_t$ are confidence$\times$visibility weights from $\mathbf{P}_t$ and $\mathbf{M}_t$, and $d_{\mathrm{SO}(3)}$ denotes the geodesic distance summed over all joints.

\parahead{Optimization details} We optimize with Adam \cite{kingma2014adam} on the sequence-shared $\boldsymbol\beta$ and all per-frame 6D rotations with a learning rate $\eta = 0.001$ for a total epochs $K = 20$. We evaluate projections with the known intrinsics $\mathbf{K}$ and translations $\mathbf{E}_t$, and compute visibility by sampling $\mathbf{M}_t$ at ground-truth keypoint locations to attenuate occluded or off-canvas points. We set $\lambda_\text{data}= 1.0, \lambda_\text{smooth} = 100.0, \alpha = 0.2, \epsilon=10^{-6}$ throughout our experiments, which yields an average runtime of \textbf{$<$1s/frame} on an NVIDIA A40 GPU.

\noindent\textbf{Why this matters?}\;
Sequence-level refinement converts strong but noisy framewise predictions into temporally stable, pixel-aligned 3D motions, which we find essential for high-quality generation (see \cref{sec:reconstruction_result} for comparisons with Animer \cite{lyu2025animer} and 4D-Fauna \cite{animal-in-motion}).




\parahead{Global Translation} The aforementioned methods operate on center-cropped videos in which the animal remains centered, allowing the model to focus solely on skeletal pose estimation. To recover global motion, we apply an off-the-shelf 3D point tracker \cite{xiao2025spatialtracker} to the original uncropped video and estimate the animal’s global translation by averaging the 3D trajectories of tracked points on the body. We then determine the scaling factor of global translation by aligning the animal size in 3D tracker coordinates and SMAL joints coordinates. Specifically, we first estimate the global translation of the animal from the tracker predictions, denoted as $\mathbf{T}_\text{tracker}$. At each time step $t$, the global translation is computed as the mean of the tracked 3D points:
\begin{equation}
\mathbf{T}^{(t)}_\text{tracker} = \frac{1}{|\mathcal{P}|} \sum_{i=1}^{|\mathcal{P}|} \mathbf{p}_i^{(t)},
\end{equation}
where $\mathcal{P} = \{\mathbf{p}_i\}$ denotes the set of tracked 3D point trajectories in world coordinate. The points are initialized by uniformly sampling a $10 \times 10$ grid on the first video frame and retaining only those located within the animal mask. The 3D trajectories ${\mathbf{p}i^{(t)}}$ are then obtained from the tracker outputs. To improve the stability of the tracker outputs, we further post-process the estimated 3D trajectories by smoothing the camera motion, aligning the reconstructed scene with a consistent ground plane, correcting residual translation drift using tracked ground points, and applying temporal smoothing to the recovered motion before estimating the animal's global translation. To align the scale of the global translation predicted by the tracker, we further rescale $\mathbf{T}_\text{tracker}$ to match the physical scale of the SMAL skeleton:

\begin{equation}
\mathbf{T}_\text{smal} = s \cdot \mathbf{T}_\text{tracker},
\end{equation}
where the scale factor $s$ is defined by matching the size of the square crop in tracker coordinates ($s_\text{tracker}$) with the head-to-tail distance of the SMAL skeleton ($s_\text{smal}$). The detailed formula to determine $s$ is left in the supplementary.

\parahead{Dataset}
We extract 3D motion for all video clips available in \cite{animal-in-motion}, yielding 29,979 motion sequences. Following their benchmark subset, we use the 230 sequence as test split, and the rest of 29,749 sequences as training split. In addition, we leverage Gemini 2.5 Flash \cite{comanici2025gemini} to infer 6 textual descriptions per video input, resulting in 179,874 textual descriptions. We show example data in the supplementary material.

\parahead{Data Validation}
Following AiM~[53], we report reconstruction metrics in main paper Table~3. Here we present human validation of motions and captions for 100 samples randomly selected from the test split. 
We consider a motion to be satisfactory if, when viewed from all angles, the motion is physically plausible with no unnatural poses, movements, or noticeable side bending of body parts, and the motion is as smooth as observed in the original video, with no perceptible jitter. Among 100 samples, 86 reconstructions are considered satisfactory. The most common failure cases arise from pure top, front, or back-view videos of moving animals, where legs are either invisible or occluded, resulting in missing leg motion in the reconstruction. Other failure cases include extremely fast animal movements or unstable camera motion in videos, which cause jitter in the reconstruction.
We evaluate captions using a scoring protocol: (0) none of the captions correctly describes the motion; (1--2) only some captions are correct; (3) all captions correctly identify the action type but contain minor errors (e.g., direction or speed); (4) all captions are correct but lack diversity; and (5) all captions are correct and cover multiple levels of detail. Among 100 samples, the average caption score is 4.62, indicating that the captions are generally accurate and diverse. None of the samples received a score of 0 or 1. 3 samples contain some captions with incorrect action (e.g., a walking video described as standing still). 7 exhibit minor errors, such as incorrect turning direction (e.g., left vs. right), often due to ambiguity in camera vs. animal perspective. The remaining 90 samples have all correct captions, among which 75 include highly diverse descriptions for the same video. We also show R-precision scores using InternVideo2 and compare with AnimalKingdom video grounding annotations. Ours shows higher scores meaning that our caption is more accurate compared to AnimalKingdom annotations. Overall, we expect our dataset to contain \textbf{86\% accurate motion reconstructions and 90\% correct caption samples.}

\subsection{Image-Conditioned Motion Diffusion}
\label{sec:image-cond-mdm}

For animals, text alone is often underspecified: species, breed, size, shape, coat, and viewpoint all affect feasible kinematics, yet are rarely captured in a short prompt (e.g., a “dog trotting” could be a tall greyhound or a stocky corgi). We therefore add an \emph{image} pathway to ground motion in visible morphology and scene context, yielding shape-aware, species-aware motion. We use images rather than videos as input, since videos would inject clip-specific motion and limit generalization, whereas images are easier to obtain and let the model learn motion priors pooled across all videos.

\parahead{Architecture}
Inspired by \cite{huang2024movein2d}, we introduce a separate branch for image conditioning on a backbone model. Building on a variation of MDM \cite{tevet2023humanmdm} with DistilBERT \cite{sanh2019distilbert} text encoder and transformer decoder \cite{vaswani2017attentiontransformer} architecture, we introduce an additional conditioning stream for image and fuse it with text and timestep conditions by simply adding them after linear projection to align feature dimension. An off-the-shelf frozen DINOv3 \cite{simeoni2025dinov3} image encoder produces a global image feature $\mathbf{z}_{\text{img}}\in\mathbb{R}^{1\times d}$; DistilBERT yields text tokens $\mathbf{z}_{\text{text}}\in\mathbb{R}^{l\times d}$, and timestep embedding $\mathbf{z}_{t} \in\mathbb{R}^{1\times d}$. The addition is performed by broadcasting and summing $\mathbf{z}_\text{img}$, $\mathbf{z}_\text{text}$, and $\mathbf{z}_t$, resulting in a combined conditioning feature $\mathbf{z}_c \in \mathbb{R}^{l \times d}$, where $l$ denotes the number of text tokens and $d$ is the feature dimension of the transformer decoder. This is then injected at every block of transformer decoder for cross-attention with the motion sequence. All other details we follow the implementation of MDM.

\parahead{Training and sampling}
We use classifier-free guidance with \emph{per-modality} dropout: independently drop text or image with probabilities \(q_{\text{text}} = 0.2\) and \(q_{\text{img}} = 0.2\) during training.
Following~\cite{tevet2023humanmdm}, we optimize the standard noise-prediction objective for diffusion. Let \(\mathbf{x}_0\in\mathbb{R}^{T\times D}\) be the motion where $D$ is the dimension of the joint representation, \(\mathbf{x}_t=\alpha_t\mathbf{x}_0+\sigma_t\boldsymbol\epsilon\) with \(\boldsymbol\epsilon\sim\mathcal{N}(0,\mathbf{I})\), and \(\mathcal{G}_\theta\) the denoiser conditioned on \(\mathbf{z}_c\):
\begin{equation}
\mathcal{L}_{\text{DDPM}}
=\mathbb{E}_{\mathbf{x}_0,c,t,\boldsymbol\epsilon}
\big\|\boldsymbol\epsilon-\mathcal{G}_\theta(\mathbf{x}_t,t,\mathbf{z}_c)\big\|_2^2 .
\end{equation}
At inference we support text-only, image-only, and text+image inputs.
Guided sampling uses CFG:
\begin{equation}
\mathcal{G}_{\mathrm{cfg}}(\mathbf{x}_t|t,c)
= \mathcal{G}(\mathbf{x}_t|t,c)
+ g\!\left[\mathcal{G}(\mathbf{x}_t|t,c) - \mathcal{G}(\mathbf{x}_t|t,\varnothing)\right],
\end{equation}
where \(c\) is the fused text–image code and \(g\) is the guidance scale.
This MDM-compatible fusion yields motions that respect the textual action while conforming to the animal morphology provided by the image.



\subsection{Applying Motions to Generated Assets}
\label{sec:assets}
To animate the generated motions on a 3D asset, we reconstruct a textured mesh from a single reference image using an image-to-3D model (Rodin~\cite{zhang2024clay}), yielding a T-pose mesh $\mathcal{M}=\{\mathbf{V},\mathbf{F}\}$.
We then retarget the generated motion to this asset in three steps:
(i) SMAL template fitting to recover the asset’s mesh-specific shape and bind-pose offsets;
(ii) skinning-weight transfer from the fitted SMAL to $\mathcal{M}$; and
(iii) linear blend skinning (LBS) to deform $\mathcal{M}$ per frame with the generated joint transforms.

\parahead{SMAL template fitting}
The generated T-pose asset is not guaranteed to match SMAL’s canonical rest pose. We therefore
estimate joint rotations that align SMAL to the asset and recover mesh-specific shape. Let
$\mathbf{V}_{\text{smal}}(\boldsymbol\beta,\boldsymbol\theta_{\text{bind}})$ be SMAL vertices posed by
per-joint rotations $\boldsymbol\theta_{\text{bind}}$. We solve
\begin{equation}
\min_{\boldsymbol\beta,\;\boldsymbol\theta_{\text{bind}}}\;
\lambda_{\text{ch}}\,D_{\text{ch}}\!\big(\mathbf{V}_{\text{smal}}(\boldsymbol\beta,\boldsymbol\theta_{\text{bind}}),\,\mathbf{V}\big)
\;+\;\lambda_{\text{e}}\,E_{\text{edge}},
\end{equation}
where $D_{\text{ch}}$ is bidirectional Chamfer and $E_{\text{edge}}$ preserves edge lengths on the
SMAL topology. This yields mesh-specific parameters $\hat{\boldsymbol\beta}$ and
per-joint “bind’’ rotations $\hat{\boldsymbol\theta}_{\text{bind}}$. The latter define the
inverse-bind corrections $\,\mathbf{B}_j{:=}\mathrm{FK}_j(\hat{\boldsymbol\beta},\hat{\boldsymbol\theta}_{\text{bind}})$
used for animation, where $\mathrm{FK}_j(\boldsymbol\beta,\boldsymbol\theta)\in SE(3)$ denotes the global
forward-kinematics transform of joint $j$ under SMAL.


\parahead{Skinning transfer}
SMAL provides LBS weights $\mathbf{W}_{\text{smal}}\!\in\!\mathbb{R}^{N_{\text{smal}}\times J}$. For each asset vertex $\mathbf{v}\in\mathbf{V}$ we compute weights by $k$-NN interpolation from fitted SMAL vertices $\tilde{\mathbf{V}}$:
\begin{equation}
\begin{aligned}
\mathbf{w}(\mathbf v) &=\sum_{i\in\mathcal N_k}\hat{\alpha}_i\,\mathbf W_{\text{smal}}[i,:], \\
\text{where } 
\hat{\alpha}_i &=\frac{\alpha_i}{\sum_{j\in\mathcal N_k}\alpha_j},\quad
\alpha_i =\frac{1}{\|\mathbf v-\tilde{\mathbf v}_i\|_2^2+\varepsilon}.
\end{aligned}
\end{equation}
We choose $k=10, \epsilon=1e-8$ in all of our experiments.

\parahead{Animation}
For each frame $t$, MDM codes $\mathbf{x}_t$ are converted to target joints $\hat{\mathbf{J}}_t$ and
we fit SMAL parameters $(\boldsymbol\beta', \boldsymbol\theta_t)$ by joint matching.
With transferred skinning weights $\mathbf{w}_j(\mathbf{v})$, we use standard LBS in homogeneous form.
Let $\tilde{\mathbf{v}}=[\mathbf{v};1]$, $\mathbf{G}_{j,t}=\mathrm{FK}_j(\boldsymbol\beta',\boldsymbol\theta_t)\!\in\!SE(3)$
be the per-frame global joint transform, and $\mathbf{B}_j=\mathrm{FK}_j(\hat{\boldsymbol\beta},\hat{\boldsymbol\theta}_{\text{bind}})$
the bind transform. Then
\begin{equation}
\tilde{\mathbf{v}}_t
= \sum_{j=1}^{J} \mathbf{w}_j(\mathbf{v})\; \big(\mathbf{G}_{j,t}\,\mathbf{B}_j^{-1}\big)\; \tilde{\mathbf{v}}.
\end{equation}
This yields textured, rigged assets driven by our synthesized animal motions.

\section{Experiments}
\label{sec:experiments}

\begin{table*}[htbp!]
    \vspace{-2em}
    \centering
    \caption{
    \textbf{Comparison with baselines on \datasetname test set.} 
    Methods are evaluated using metrics from \cite{guo2022t2m}, with top results in \textcolor{tabfirst}{\textbf{red}} (best) and \textcolor{tabsecond}{\textbf{blue}} (second-best).
    We report each metric's average and 95\% confidence interval, based on 10 evaluations.
    }
    \vspace{-0.5em}
    \resizebox{\textwidth}{!}{

    \begin{tabular}{l c c c c c c c}
    
    \toprule
    \multirow{2}{*}{Methods}  & \multicolumn{3}{c}{R-Precision $\uparrow$} & \multirow{2}{*}{FID $\downarrow$} & \multirow{2}{*}{MM-Dist $\downarrow$} & \multirow{2}{*}{Diversity $\uparrow$} & \multirow{2}{*}{MModality $\uparrow$}\\

    \cmidrule{2-4}
    ~ & Top-1 & Top-2 & Top-3 \\
    
    \midrule

        Ground-Truth    &
        \et{ 0.113 }{ .002 }      & 
        \et{ 0.206 }{ .003 }      & 
        \et{ 0.281 }{ .003 }      & 
        \et{ 0.004 }{ .001 }      & 
        \et{ 5.517 }{ .008}      & 
        \et{ 3.934 }{ .076 }      & 
        -                       \\

    \midrule
        AniMo (\textit{AniMo4D \cite{wang2025animo} data}) & 
        \et{ 0.029 }{ .001 }                        & 
        \et{ 0.059 }{ .001 }                        & 
        \et{ 0.089 }{ .001 }                        & 
        \et{ 30.043 }{ .086 }                        & 
        \et{ 8.564 }{ .007 }                        & 
        \et{ 3.713 }{ .055 }                          & 
        \et{ 3.723 }{ .009 }                          \\

        AniMo (\textit{\datasetname data}) & 
        \et{ 0.029 }{ .001 }                        & 
        \et{ 0.058 }{ .001 }                        & 
        \et{ 0.088 }{ .002 }                        & 
        \et{ 30.516 }{ .072 }                        & 
        \et{ 8.659 }{ .007 }                        & 
        \et{ 3.800 }{ .060 }                          & 
        \et{ 3.72 0 }{ .008 }                          \\

    \midrule
        \methodname (ours, \textit{text})     &
        \etsecond{ 0.032 }{ .001 } & 
        \etsecond{ 0.065 }{ .001 } & 
        \etsecond{ 0.097 }{ .001 } & 
        \etsecond{ 11.889 }{ .024 } & 
        \etsecond{ 7.007 }{ .003 } & 
        \etfirst{ 4.693 }{ .071 } & 
        \etfirst{ 4.609 }{ .011 }\\

        \methodname (ours, \textit{image}) & 
        - & - & - &
        \et{25.089}{.055} &
        - &
        \et{1.761}{.034} &
        \et{5.804}{.029}
        \\
        
        \methodname (ours, \textit{text + image})       &
        \etfirst{ 0.043 }{ .001} & 
        \etfirst{ 0.087 }{ .001 } & 
        \etfirst{ 0.130 }{ .002 } & 
        \etfirst{ 6.248 }{ .014 } & 
        \etfirst{ 6.218 }{ .004 } & 
        \etsecond{ 4.319 }{ .093 } & 
        \etsecond{ 4.272 }{ .017 }\\

    \bottomrule
    \end{tabular}
    }
    \vspace{-1em}
    \label{tab:motion_ours}
\end{table*}

\begin{table*}[htbp!]
\vspace{-2em}
\centering
\small
\caption{
\textbf{Comparison with baselines on AnimalML3D \cite{yang2023omnimotiongpt} data.}
Best results are shown in \textcolor{tabfirst}{\textbf{red}}.
}
\vspace{-0.5em}
\resizebox{\textwidth}{!}{
\begin{tabular}{l c c c c c c c}
    \toprule
    \multirow{2}{*}{Methods}  & \multicolumn{3}{c}{R-Precision $\uparrow$} & \multirow{2}{*}{FID $\downarrow$} & \multirow{2}{*}{MM-Dist $\downarrow$} & \multirow{2}{*}{Diversity $\uparrow$} & \multirow{2}{*}{MModality $\uparrow$}\\

    \cmidrule{2-4}
    ~ & Top-1 & Top-2 & Top-3 \\
    
    \midrule

        Ground-Truth    &
        \et{ 0.964 }{ .028 }      & 
        \et{ 1.000 }{ .000 }      & 
        \et{ 1.000 }{ .000 }      & 
        \et{ 0.235 }{ .080 }      & 
        \et{ 1.709 }{ .084 }      & 
        \et{ 11.778 }{ .874 }      & 
        -                       \\

    \midrule
        AniMo (\textit{AniMo4D \cite{wang2025animo} data}) & 
        \et{ 0.031 }{ .001 }                        & 
        \et{ 0.062 }{ .001 }                        & 
        \et{ 0.093 }{ .001 }                        & 
        \et{ 150.671 }{ .0.913 }                        & 
        \et{ 17.171 }{ .052 }                        & 
        \et{ 5.152 }{ .126 }                          & 
        \etfirst{ 5.159 }{ .023 }                          \\

        AniMo (\textit{\datasetname data}) & 
        \et{ 0.031 }{ .001 }                        & 
        \et{ 0.061 }{ .001 }                        & 
        \et{ 0.093 }{ .001 }                        & 
        \et{ 155.687 }{ 0.739 }                        & 
        \et{ 12.334 }{ .069 }                        & 
        \et{ 4.571 }{ .077 }                          & 
        \et{ 4.579 }{ .015 }                          \\
        
    \midrule
        \methodname (ours)     &
        \etfirst{ 0.039 }{ .001 } & 
        \etfirst{ 0.076 }{ .001 } & 
        \etfirst{ 0.111 }{ .001 } & 
        \etfirst{ 138.145 }{ .725 } & 
        \etfirst{ 12.128 }{ .003 } & 
        \etfirst{ 5.152 }{ .057 } & 
        \et{ 5.126 }{ .020 }\\

    \bottomrule
    \end{tabular}
}
\vspace{-2em}
\label{tab:motion_animalml3d}
\end{table*}

\paragraph{Baselines.}  
\textbf{Motion generation:} We compare against against AniMo \cite{wang2025animo}, which is, to our knowledge, the only publicly available method tailored for \emph{text-driven} animal motion generation. We compare with AniMo trained on ours dataset and on their original AniMo4D dataset. We report our text-only and \emph{text+image} setting to show that visual conditioning further stabilizes pose and style.
\textbf{Mesh animation:} We qualitative compare with Puppeteer \cite{song2025puppeteer}, a recent pipeline that also accepts text and image inputs to produce animated animal meshes. For a fair comparison, we use the same text, image, and generated T-posed mesh inputs for both methods, and follow Puppeteer’s setup by employing Kling AI \cite{KlingAI_global} as the video generation module.

\paragraph{Datasets.}
For motion evaluation, we evaluate on two animal motion datasets. We first show results on \datasetname, which is our reconstructed motion dataset using videos from AiM \cite{animal-in-motion}. We use their cleaned benchmark data as test split, which has 230 motions, and we exclude them from training set. The same benchmark is also used for quantitatively evaluate motion reconstruction.
Following \cite{wang2025animo}, we also show results on AnimalML3D \cite{yang2023omnimotiongpt}, which is an external out-of-distribution test set that none of the methods has trained on. AnimalML3D contains 1,260 hand-crafted motions curated from DeformingThings4D \cite{li20214dcomplete}.

\paragraph{Metrics.}
For motion generation, we follow standard text-to-motion evaluation \cite{guo2022t2m}:
(1) \textbf{R-Precision}: text-motion retrieval accuracy in top-k accuracies;
(2) \textbf{FID}: Fréchet distance between generated and real motion distributions;
(3) \textbf{MM-Dist}: distance in a shared text-motion latent space;
(4) \textbf{Diversity}: average pairwise distance between independently sampled motions;
(5) \textbf{Multimodality}: variance among multiple motions generated from the same text.

For motion reconstruction, we follow the metrics described in \cite{animal-in-motion}:
For evaluating 4D animal reconstruction, we use the following metrics: 
(1) \textbf{Silhouette IoU}: IoU between the rendered silhouette and the ground-truth mask; 
(2) \textbf{PCK}: percentage of projected keypoints that fall within a normalized distance threshold; 
(3) \textbf{KT}: PCK error after a 2D-to-3D-to-2D re-projection to a novel view, used as a proxy for 3D shape consistency; 
(4) \textbf{MPJVE}: average error between predicted and ground-truth joint velocities in projected pixel space to evaluate temporal motion.

\subsection{Quantitative Results on Motion Generation}
\label{motion_result}

For each text in the evaluation dataset, we generate 10 motion samples and repeat the evaluation over 10 trials. We report the mean and 95\% confidence interval across trials, as shown in \cref{tab:motion_ours}. We evaluate four configurations: the baseline model trained on its original data AniMo4D \cite{wang2025animo}, the baseline retrained on our dataset, and our model in two variants—(1) trained with both image and text conditioning, and (2) trained and evaluated with the image branch disabled (text-only).

As shown in \cref{tab:motion_ours}, both of our variants outperform the baseline across all metrics. Retraining the baseline on our dataset already leads to a improvement, demonstrating the quality of text and motion quality of our newly constructed dataset. The text-only version of our model further surpasses the baseline trained on the same data, indicating that our architecture and method models the motion more effectively. Incorporating image conditioning yields additional gains on most metrics, achieving the best overall fidelity and text–motion alignment, producing motions that are both semantically coherent and visually plausible. Together, these results validate the effectiveness of both our dataset and our model design, highlighting the benefit of integrating textual and visual signals for animal motion generation.

To further demonstrate the generality and quality of our dataset, we evaluate on an external test-only dataset, AnimalML3D \cite{yang2023omnimotiongpt}, where neither our models nor the baselines have been trained. We compare our methods trained on our dataset against AniMo trained on AniMo4D \cite{wang2025animo}. As shown in \cref{tab:motion_animalml3d}, our models outperform the baseline across all metrics, confirming that training on our dataset leads to better generalization and higher-quality motion modeling, even on unseen data.

\subsection{Quantitative Results on Reconstruction}
\label{sec:reconstruction_result}
\begin{table*}[htbp!]
    \vspace{-2.2em}
    \centering
    \small
    \caption{\textbf{Benchmark comparison of 4D animal reconstruction methods.} Our method maintains nearly real-time processing speeds comparable to feed-forward approaches \cite{lyu2025animer,li2024learning3dfauna}, while achieving improved performance across all evaluation metrics.}
    \vspace{-0.5em}
    \resizebox{\textwidth}{!}{
    \begin{tabular}{lccccccc}
        \toprule
         Method & IoU $\uparrow$ & PCK@0.1 $\uparrow$ & PCK@0.05 $\uparrow$ & KT-PCK@0.1 $\uparrow$ & KT-PCK@0.05 $\uparrow$ & MPJVE $\downarrow$ & Time $\downarrow$\\
        \midrule
        SMALify \cite{biggs2020leftsmalify} & 0.867 & 0.954 & 0.787 & 0.623 & 0.372 & 0.023 & $\sim$ 30\text{s/frame} \\
        4D-Fauna \cite{animal-in-motion} & 0.814 & 0.664 & 0.317 & 0.418 & 0.193 & 0.044 & $\sim$ 30\text{s/frame}  \\
        \midrule
        AniMer \cite{lyu2025animer} & 0.677 & 0.537 & 0.199 & 0.566 & 0.256 & 0.038 & $<$ 1\text{s/frame}\\
        3D-Fauna \cite{li2024learning3dfauna} & 0.670 & 0.470 & 0.177 & 0.329 & 0.130 & 0.058 & $<$ 1\text{s/frame}\\
        \midrule
        \methodname (ours) & 0.698 & 0.751 & 0.485 & 0.614 & 0.332 & 0.037 & $<$ 1\text{s/frame} \\
        \bottomrule
    \end{tabular}
    }
    \vspace{-2.2em}
    \label{tab:reconstruction}
\end{table*}

\begin{figure*}[!htpb]
\vspace{-1em}
    \centering
    \includegraphics[width=0.95\linewidth]{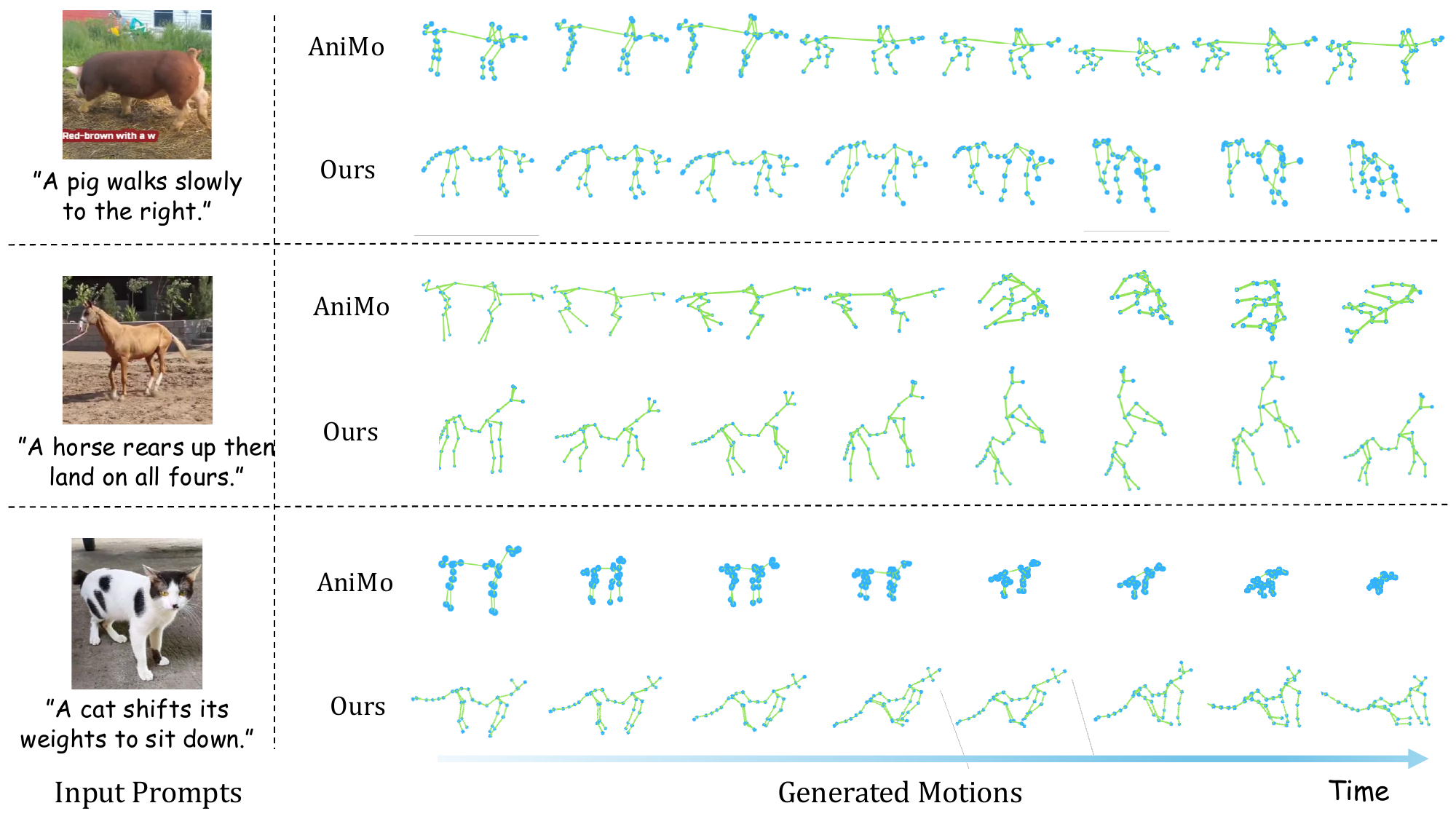}
    \vspace{-1em}
    \caption{\textbf{Visual comparison of generated skeletal motion with the baseline.} The left columns show the input text and image. AniMo uses only text, while our method conditions on both text and image. AniMo exhibits failures such as motions that do not follow the prompt and inconsistent skeleton shapes, whereas our method produces more realistic motion sequences.}
    \vspace{-1.5em}
    \label{fig:skeleton}
\end{figure*}

\begin{figure*}[htpb]
    \centering
    \includegraphics[width=0.87\linewidth]{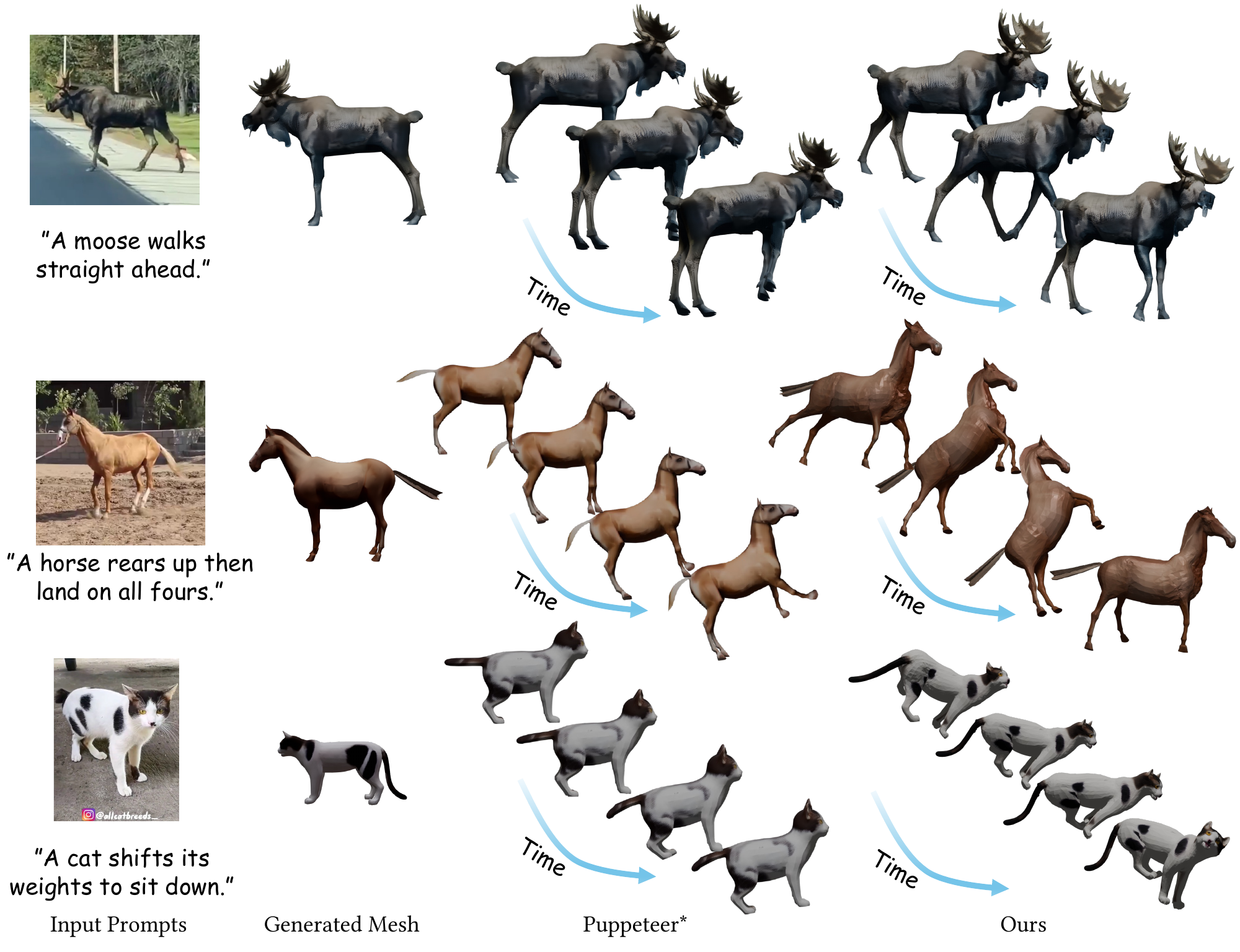}
    \vspace{-0.5em}
    \caption{
    \textbf{Visual comparison of generated mesh animation with the baseline.} The left columns show the input text and image, which are used for both pipelines. Puppeteer often produces little or no motion on the input mesh, whereas our method generates realistic movements that follow the text prompt.
    }
    \vspace{-1em}
    \label{fig:puppeteer}
\end{figure*}

Following \cite{animal-in-motion}, we evaluate our motion reconstruction method on their benchmark, with results shown in \cref{tab:reconstruction}. Among the baselines, AniMer \cite{lyu2025animer} and 3D-Fauna \cite{li2024learning3dfauna} are feed-forward methods that operate in near real time, while SMALify \cite{biggs2020leftsmalify} and 4D-Fauna \cite{animal-in-motion} are optimization-based approaches that require significantly longer computation time to fit each sequence. Although our method includes a post-optimization step after feed-forward inference, it still achieves processing speeds comparable to feed-forward baselines, while simultaneously improving both projection-based and 3D-aware metrics. These results demonstrate that our reconstruction method is both accurate and efficient, making it suitable for large-scale data processing.

\subsection{Qualitative Results}
We qualitatively compare our motion generation results with AniMo \cite{wang2025animo} as the baseline and observe several notable failure cases in the baseline outputs. As shown in \cref{fig:skeleton}, in the first example, the baseline fails to follow the text input and produces an unrealistic walking motion, while our method successfully generates a coherent walking sequence that naturally turns left. In the second example, the baseline misinterprets the text prompt “rearing up,” generating a lying-down motion instead, whereas our model correctly produces a smooth rearing-up and landing motion for the horse. A particularly critical failure of the baseline is its inconsistent skeleton structure across frames: as illustrated in the third example, although the cat roughly follows the “sitting” prompt, its skeleton collapses and shrinks to an unrealistic size, making the motion unusable for downstream tasks such as mesh rigging. In contrast, our method maintains consistent bone lengths and body proportions, resulting in stable and anatomically plausible motions that better align with the input text descriptions.

Qualitative results comparing with Puppeteer \cite{song2025puppeteer} show that our method produces more accurate and natural motions. While Puppeteer relies on optimizing a rigged mesh to match frames generated by a video generation model, this indirect approach often fails when the generated videos are inconsistent or physically implausible. In particular, we observe frequent failure cases where the generated video does not follow the input text, deviates from the initial frame, or exhibits abrupt shot changes and inconsistent animal appearance and shapes. These issues make motion extraction unreliable.

In contrast, our method learns motion directly from real-world animal videos, reconstructing temporally smooth 3D joint trajectories that generalize robustly to new inputs. As a result, rigging and animating a mesh using our generated motion is consistently more stable and faithful to the input text and image. This demonstrates that learning motion from real video data provides a more reliable and physically grounded foundation than relying on motion cues extracted from generated videos, validating the effectiveness of our pipeline design.

\subsection{Ablation Studies}

We conduct an ablation study to evaluate the effect of image conditioning in our motion generation model. As shown in \cref{tab:motion_ours}, we compare models trained with and without the image-conditioning branch while keeping all other training configurations identical.

Adding image conditioning leads to consistent improvements across key perceptual and alignment metrics: R-Precision, FID, and MM-Distance all show notable gains, with FID exhibiting the largest improvement. This indicates that integrating visual cues helps the model generate more realistic and visually coherent motions that align better with both text and ground-truth motion distributions. The improvement in R-Precision and MM-Distance also confirms that image conditioning enhances semantic alignment between text and motion, suggesting that visual features provide an effective bridge between the two modalities.
On the other hand, Diversity and Multimodality scores remain comparable but showing a marginal decrease when image features are used. We attribute this to the nature of visual conditioning, which introduces stronger spatial and appearance constraints that guide the motion synthesis process more tightly toward visually plausible outcomes, potentially reducing motion variation.


\section{Conclusion}

We introduce \methodname, a framework for learning and generating realistic 3D animal motion from in-the-wild videos. To address the challenge of data scarcity, we construct \datasetname, the first large-scale animal dataset containing aligned video, text, and 3D motion tuples. Building on this dataset, we develop an animal motion generation model that conditions on both text and image. Experiments show that our model achieves state-of-the-art performance on both in-distribution and external test sets. Finally, we present an animation pipeline that rigs the generated motions onto 3D meshes. Together, our reconstruction method and dataset, generative model, and rigging pipeline form a unified solution for data-driven animal motion modeling and animation. We believe this work opens new opportunities for biomechanics, behavioral analysis, and realistic character animation for media. All code and datasets will be released.

\section*{Acknowledgments}
This work is in part supported by NSF RI \#2211258 and \#2338203, ONR MURI N00014-22-1-2740, and ONR MURI N00014-24-1-2748.

\clearpage

%
%
\bibliographystyle{splncs04}
\bibliography{main}

\clearpage
\setcounter{page}{1}
\setcounter{section}{0}
{
\centering
\Large
\textbf{\methodname: Animal Motion Generation from In-the-Wild Video}\\[0.5em]
{\Large -- Supplementary Material --} \\
}

\section{Dataset Details}
\begin{figure*}[!htpb]
    \centering
    \includegraphics[width=\linewidth]{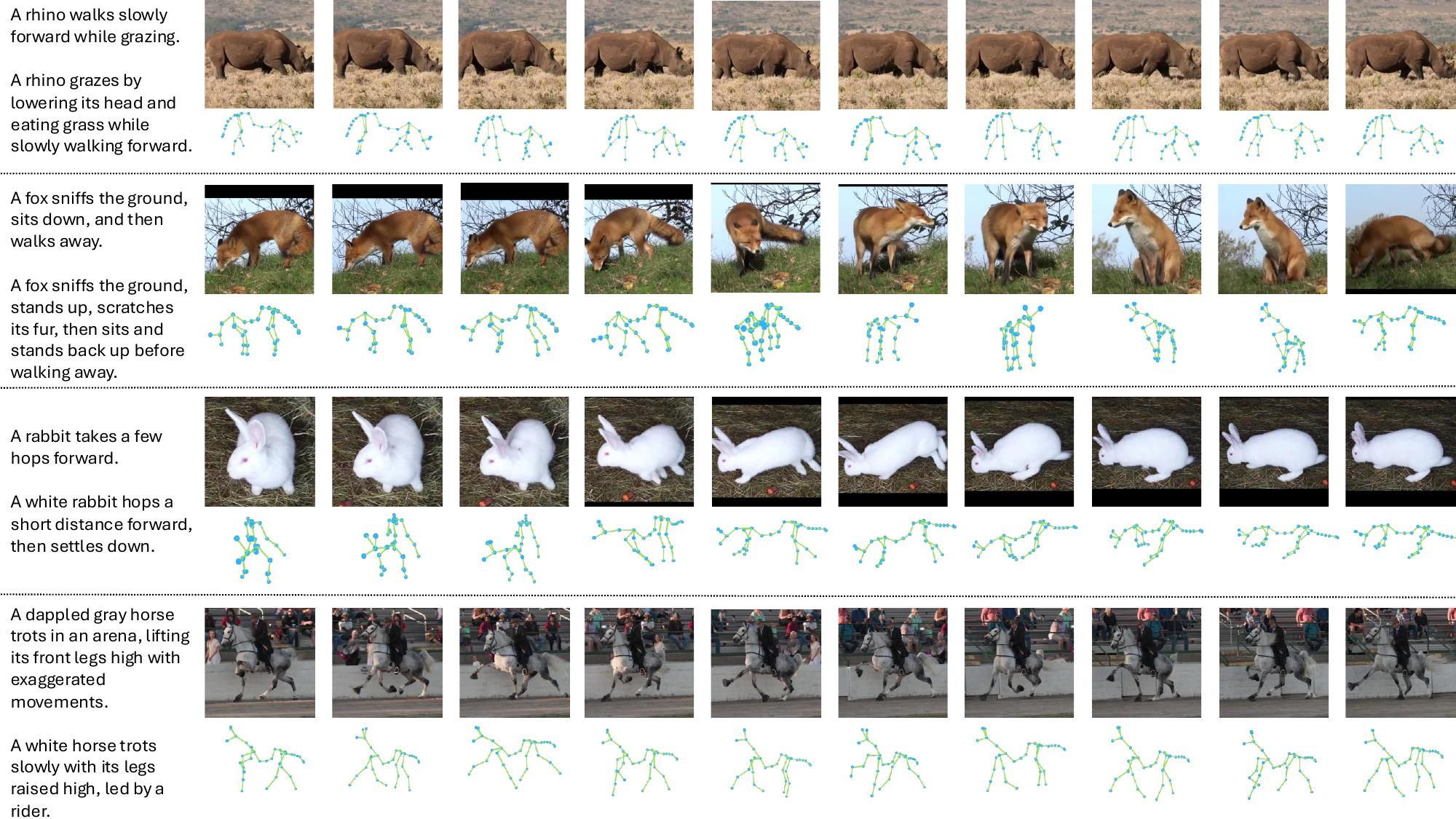}
    \caption{\textbf{Visualization of animal articulation of data examples.} For each row, the left shows two textual descriptions of the motion, followed by the video frames and corresponding reconstructed 3D motion.}
    \vspace{-3mm}
    \label{fig:data_example}
\end{figure*}

\begin{figure*}[!htpb]
    \centering
    \includegraphics[width=\linewidth]{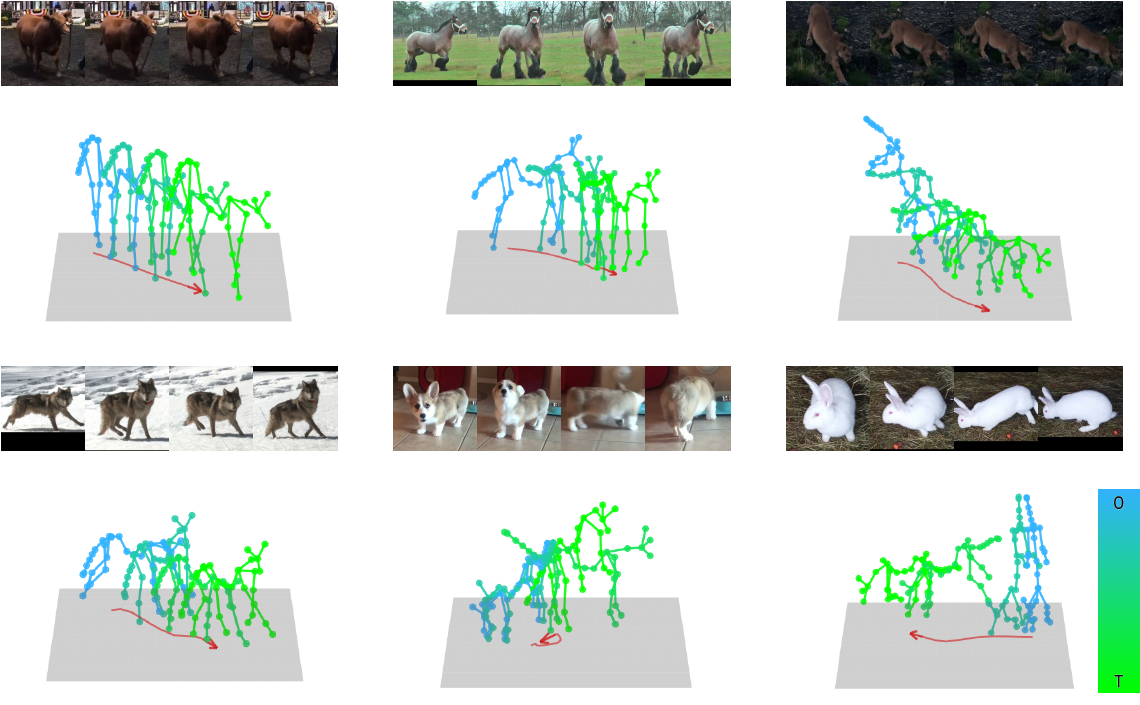}
    \caption{\textbf{Visualization of global translations of data examples.} Each data example shows the trajectory of the root joint projected onto the floor, visualizing the global translation of the reconstructed motion.}
    \vspace{-3mm}
    \label{fig:data_example_global}
\end{figure*}

Since our dataset build upon AiM \cite{animal-in-motion}, our dataset have the same number of motion data samples, with 29,979 motions, where 230 of them, 10 for each of the 23 categories, are used for test set, and the remaining 29,749 are used as training set. We compare the number of motions across existing animal-motion datasets in \cref{tab:dataset_comparison}. Unlike prior datasets, which rely on motions manually crafted by human designers, ours is derived directly from real in-the-wild videos. Some data examples are shown in \cref{fig:data_example}, which visualizes the articulation of the animal, and in \cref{fig:data_example_global}, which visualizes the global translation of the reconstructed motion.

\section{Caption Validation}

We further validate the generated captions using retrieval based metrics. Specifically, we evaluate the R Precision score at top 1, top 2, and top 3, and compare the results with the video grounding subset of AnimalKingdom \cite{animalkingdom}. The results are shown in \cref{tab:caption_retrieval}.

\begin{table}[t]
\centering
\caption{Comparison of video grounding performance on different datasets.}
\label{tab:caption_retrieval}
\resizebox{0.5\linewidth}{!}{
    \begin{tabular}{lccc}
    \toprule
    Dataset & R1 & R2 & R3 \\
    \midrule
    AnimalKingdom (video grounding) & 0.10 & 0.20 & 0.28 \\
    \midrule
    \rowcolor[gray]{.95} \textbf{AiM3D (Ours)} & \textbf{0.10} & \textbf{0.23} & \textbf{0.35} \\
    \bottomrule
    \end{tabular}
}
\end{table}

\section{Dataset Distribution}
\begin{figure}[t]
    \centering
    \includegraphics[width=0.8\linewidth]{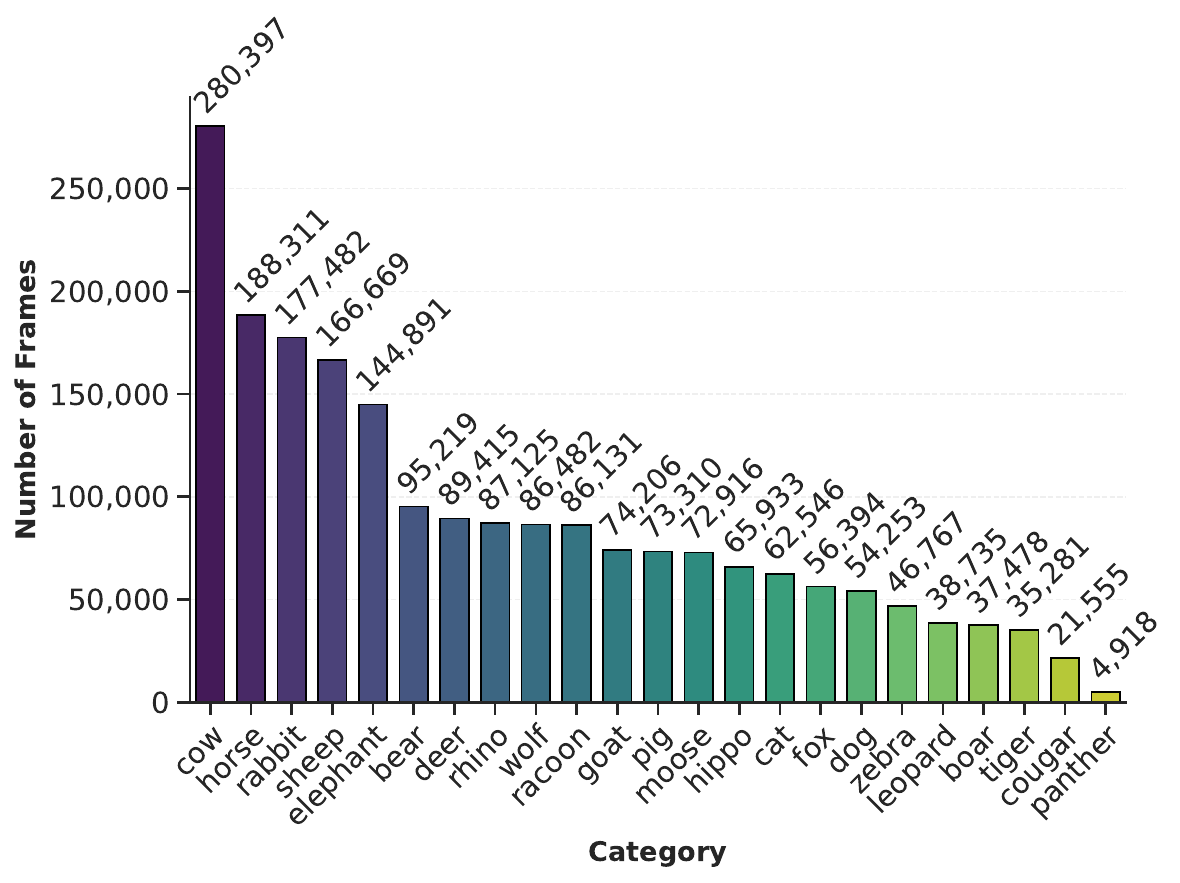}
    \caption{Number of frames for each animal category.}
    \label{fig:data_category_frame_stats}
\end{figure}

\begin{figure}[t]
    \centering
    \includegraphics[width=0.8\linewidth]{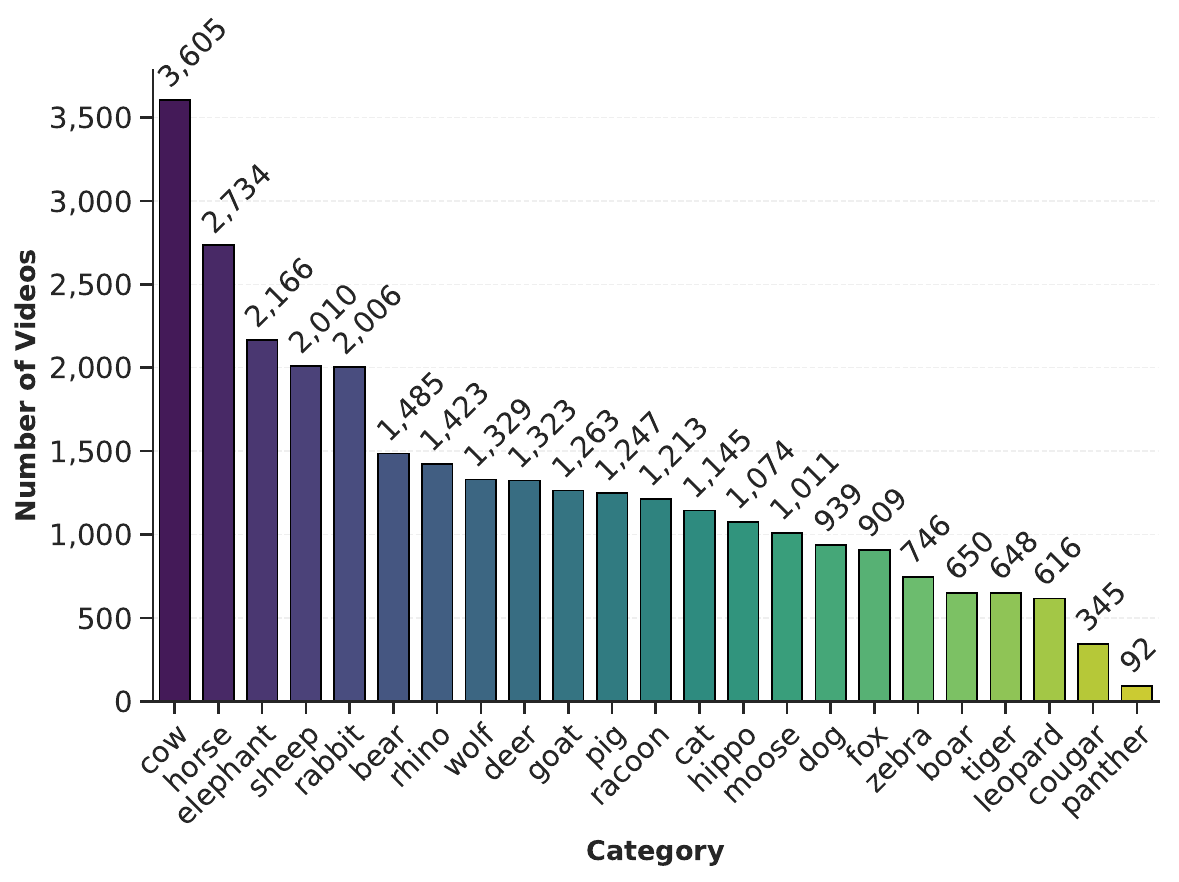}
    \caption{Number of videos for each animal category.}
    \label{fig:data_category_video_stats}
\end{figure}

\begin{figure}[t]
    \centering
    \includegraphics[width=0.8\linewidth]{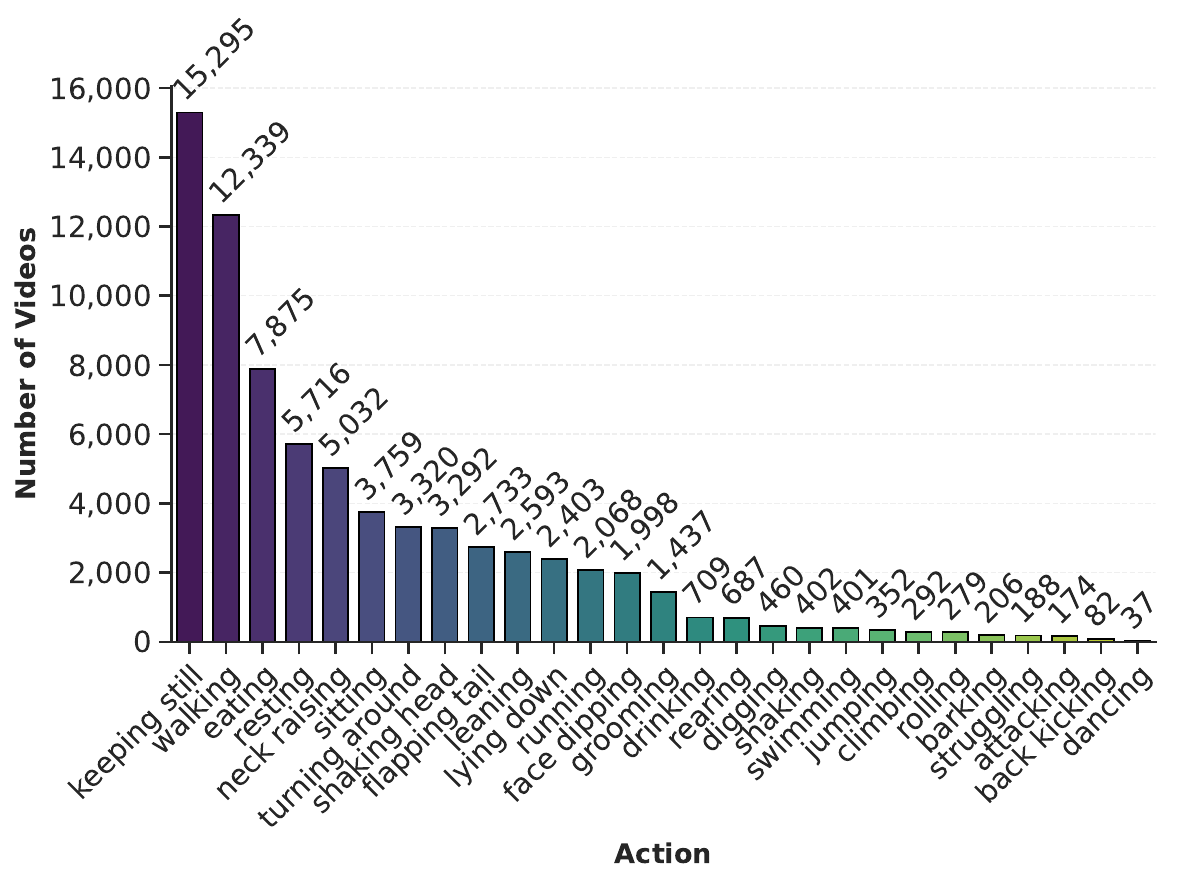}
    \caption{Distribution of motion types across the full dataset. Each video is assigned one to three motion labels.}
    \label{fig:data_motion_stats}
\end{figure}

Since our dataset is built upon the AiM dataset \cite{animal-in-motion}, the distributions of animal categories and motion types follow those of the original dataset. For reference, these distributions are shown in \cref{fig:data_category_frame_stats,fig:data_category_video_stats,fig:data_motion_stats}.

\section{Comparison with Existing Dataset}

To the best of our knowledge, \datasetname is the only large scale animal dataset that simultaneously provides video, caption, and motion annotations. As shown in \cref{tab:dataset_comparison}, existing datasets typically focus on a single modality, such as video collections or motion datasets, or text and human crafted motion. In contrast, our dataset unifies these modalities by providing aligned in-the-wild video, textual descriptions, and reconstructed motion sequences, enabling multimodal learning for animal motion understanding and generation.

\begin{table}[h]
\caption{\textbf{Comprehensive comparison with existing animal datasets across modalities.}}
\centering
\begin{tabular}{||l|c|c|c|c||} 
 \hline
 Dataset & \#Video & \#Caption & \#Motion & Motion Source \\ 
 \hline\hline
 MammalNet \cite{chen_mammalnet_2023} & 18,346 & NA & NA & NA \\
 \hline
 AnimalKingdom \cite{animalkingdom} & 4,301 & 18,744 & NA & NA \\
 \hline
 Truebones Zoo \cite{truebones} & NA & NA & 1,219 & Human-crafted \\
 \hline
 AnimalML3D \cite{yang2023omnimotiongpt} & NA & 3,720 & 1,240 & Human-crafted \\ 
 \hline
 AniMo4D \cite{wang2025animo} & NA & 185,435 & 78,149 & Human-crafted \\
 \hline
 PFERD \cite{li2024posespferd} & 460 & NA & 46 & Motion capture \\
 \hline
 AiM \cite{animal-in-motion} & 29,979 & NA & NA & NA \\
 \hline
 \textbf{\datasetname (Ours)} & \textbf{29,979} & \textbf{179,874} & \textbf{29,979} & \textbf{In-the-wild Video} \\
 \hline
\end{tabular}

\label{tab:dataset_comparison}
\end{table}

\section{VLM Prompts}
We leverage Gemini 2.5 Flash as the VLM to infer textual descriptions for video data in AiM dataset \cite{animal-in-motion}. For each video input to VLM, we use 6 different prompts to generate 6 descriptions with variations. We use a universal system message:

\begin{tcolorbox}[colback=gray!5!white, colframe=gray!75!black]
\textit{You are annotating a text-to-motion generation dataset. You will be given a video clip of animal motion, and you will be asked to write concise motion captions suitable for training a text-to-motion generation model. }
\end{tcolorbox}

Then we apply the following prompts independently to the same input video, where \{\textit{animal}\} is replaced by specific animal categories.

\begin{tcolorbox}[colback=gray!5!white, colframe=gray!75!black]
\textit{
    Describe the \{animal\}'s motion in a single short sentence.
}
\end{tcolorbox}
\begin{tcolorbox}[colback=gray!5!white, colframe=gray!75!black]
\textit{
The \{animal\} is doing what kind of motion in the video? Answer in one short sentence.
}
\end{tcolorbox}

\begin{tcolorbox}[colback=gray!5!white, colframe=gray!75!black]
\textit{
Use simple sentences to describe the \{animal\}'s motion and global movement in the video.
}
\end{tcolorbox}

\begin{tcolorbox}[colback=gray!5!white, colframe=gray!75!black]
\textit{What is the main action of the \{animal\} in the video? Be specific about **speed, style, and sequence of actions**.
}
\end{tcolorbox}

\begin{tcolorbox}[colback=gray!5!white, colframe=gray!75!black]
\textit{Describe the \{animal\}'s motion in this video. Focus on the detailed motion.}
\end{tcolorbox}

\begin{tcolorbox}[colback=gray!5!white, colframe=gray!75!black]
\textit{Write a short caption of the \{animal\} motion in this video that describes the **motion type** and **global movement path**.
}
\end{tcolorbox}

We append each prompt input with an in-context example for style reference, where \{\textit{example}\} is a text description randomly selected from AniMo4D \cite{wang2025animo} and AnimalML3D \cite{yang2023omnimotiongpt} data.

\begin{tcolorbox}[colback=gray!5!white, colframe=gray!75!black]
\textit{
Following the style of this example: \{example\}
}
\end{tcolorbox}

\section{Global Translation Scaling Details}

\begin{equation}
s = \frac{s_\text{smal}}{s_\text{tracker}}
= \frac{s_\text{smal}}{
\frac{1}{2}\left(\frac{c_\text{crop}^{(0)}}{f_x^{(0)}}+\frac{c_\text{crop}^{(0)}}{f_y^{(0)}}\right)
\cdot \frac{1}{|\mathcal{P}|}\sum_{i=1}^{|\mathcal{P}|} d_i^{(0)}
},
\end{equation}

in which $c_\text{crop}^{(0)}$ denotes the crop box size in the original video frame, $f_x^{(0)}$ and $f_y^{(0)}$ are the camera focal lengths, and $d_i^{(0)}$ is the depth of the $i$-th tracked point, all evaluated at the first frame of the video.

\section{Motion Reconstruction Details}
\begin{figure*}[!htpb]
    \centering
    \includegraphics[width=\linewidth]{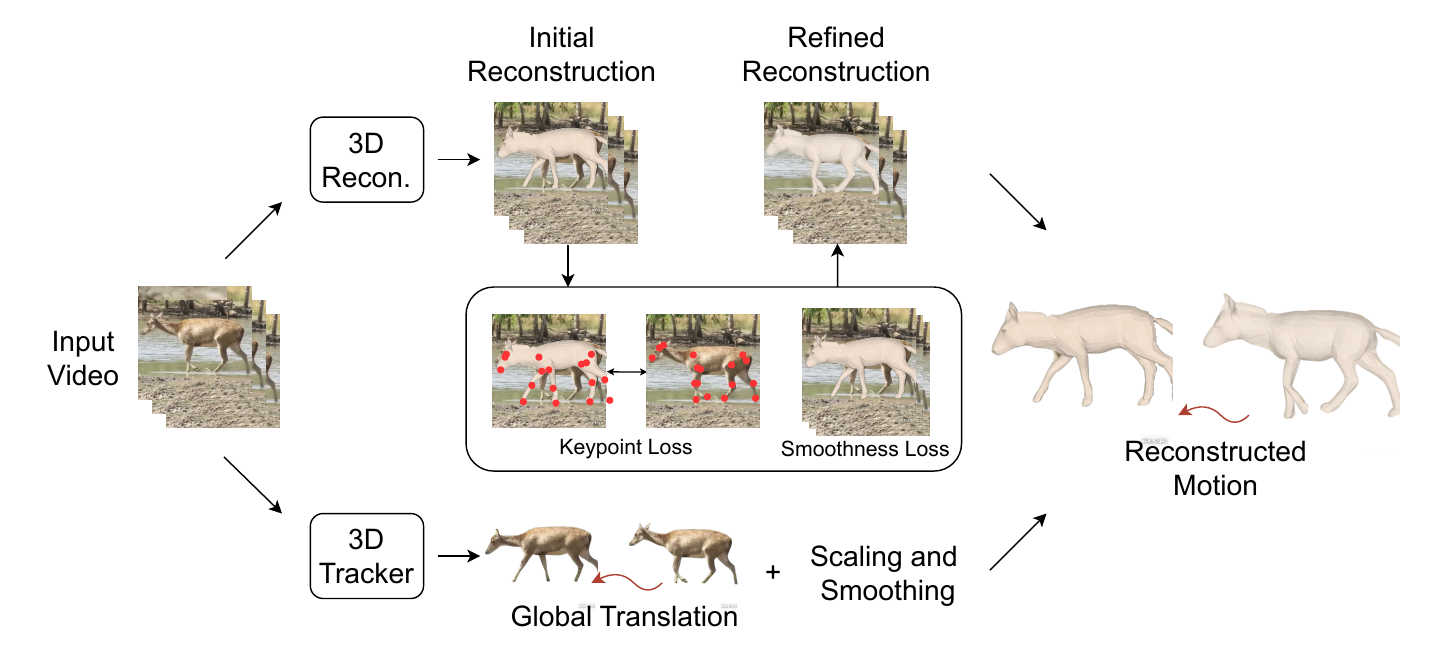}
    \caption{\textbf{Overview of motion reconstruction method.} We leverage off-the-shelf 3D quadruped reconstruction method and 3D tracking method to infer articulation and global translation separately. We combine them to obtained the final motion reconstruction as our motion data. }
    \vspace{-3mm}
    \label{fig:recon_method}
\end{figure*}

We show an overview figure of our motion reconstruction method in \cref{fig:recon_method}.

\begin{figure*}[!htpb]
    \centering
    \includegraphics[width=\linewidth]{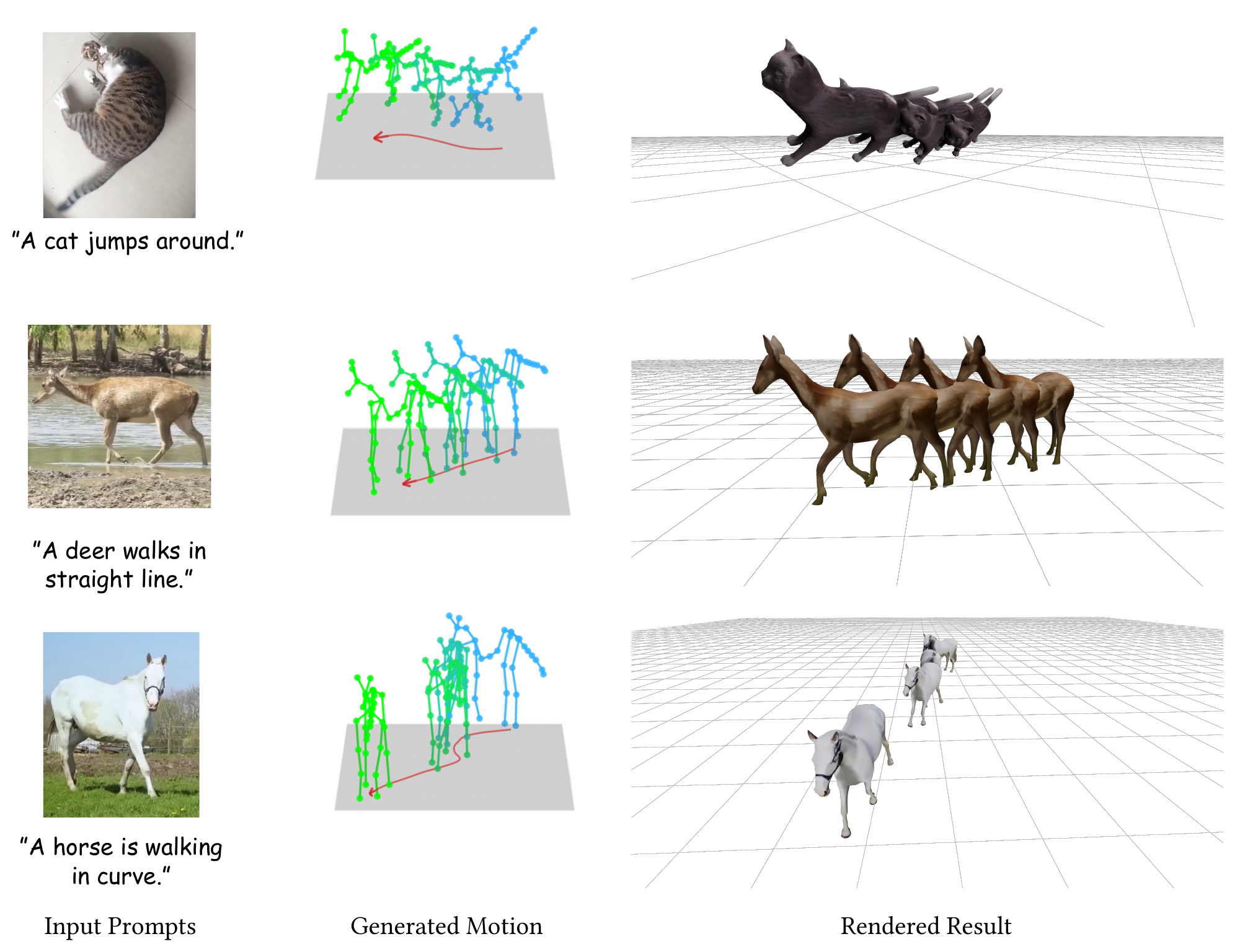}
    \caption{\textbf{Additional Results.} Additional examples showing more noticeable global motion. }
    \vspace{-3mm}
    \label{fig:additional}
\end{figure*}
\section{Additional Results}
We provide additional results with more evident global motion in \cref{fig:additional}. Readers are encouraged to refer to our project page for interactive visualizations.

\section{Image Condition}

\begin{figure*}[!htpb]
    \centering
    \includegraphics[width=\linewidth]{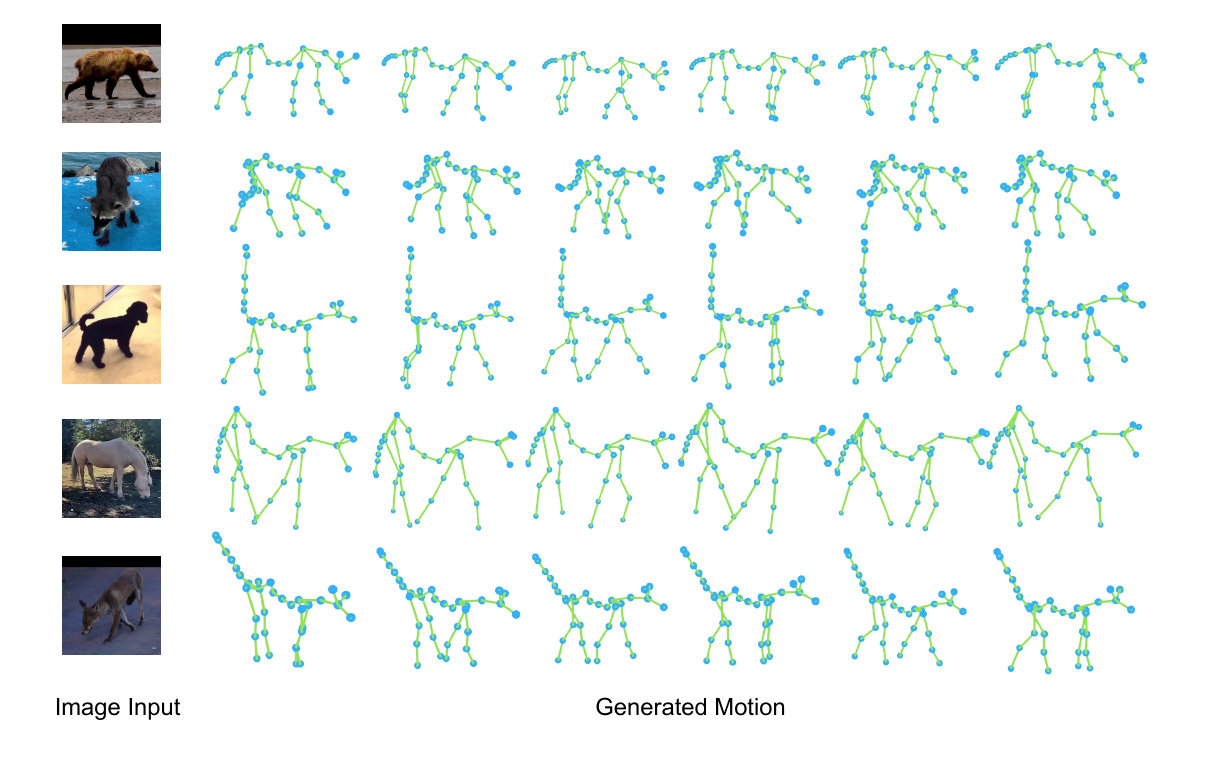}
    \caption{\textbf{Effect of Image Conditioning.} The generated motions use the same input text prompt, \textit{An animal is walking}. The results show appropriate skeleton variations and gait patterns conditioned on the input image. }
    \vspace{-3mm}
    \label{fig:image_condition}
\end{figure*}

We demonstrate the effect of image conditioning on generated motion in \cref{fig:image_condition}. Specifically, the image condition influences both the gait and the skeletal morphology. When using the same text prompt, \textit{An animal is walking}, the generated results vary according to the input image condition.

\section{Limitaion}


Although image conditioning has proven effective, our generation pipeline currently lacks animal specific structural priors. The same model is used across species of vastly different scales and skeletal proportions, for example elephant versus rabbit, which may limit species specific realism. Future designs could incorporate adaptive skeleton representations or category aware conditioning to better account for diverse anatomies.

Moreover, our current framework does not explicitly model tail motion. Many source videos either omit the tail due to occlusion or truncation, and the pseudo ground truth keypoint annotations do not include tail joints. Consequently, the reconstructed skeleton ignores tail placement and dynamics, leading to inconsistent or arbitrary tail motion in both reconstructed and generated sequences. Since the tail can convey important semantic and behavioral information, particularly for species such as cats, future work could focus on constructing datasets with reliable tail annotations or incorporating dedicated tail modeling modules.

In addition, because motion is reconstructed from monocular video, it inevitably suffers from depth ambiguity and occasional reconstruction failures. These errors can accumulate and result in physically implausible artifacts, such as foot sliding or floating. While our current pipeline does not explicitly enforce physical constraints, future approaches could incorporate stronger physics aware post processing, reinforcement learning based refinement, or knowledge transfer from motion capture data of domesticated animals such as cats and dogs to improve physical realism.

Furthermore, since reconstruction is performed from single view video, heavy occlusion or limited viewpoints can significantly degrade motion quality. In cases where the animal is captured from a purely frontal or top view, or when limb motion is largely occluded, the model must infer leg trajectories without sufficient visual evidence. This ambiguity can lead to incorrect limb estimation and physically implausible motion in the reconstructed data, which may subsequently affect generation quality. Future work could address this limitation by incorporating multi view supervision, stronger temporal priors, or explicit visibility aware modeling.

Finally, our auto rigging step involves a post optimization process to fit a predefined skeleton to generated joint sequences, which can introduce misalignment and minor reconstruction errors. Future approaches could mitigate these issues by employing learned mesh rigging and retargeting methods, or by directly predicting SMAL parameters to eliminate the need for post hoc optimization.

\end{document}